\documentclass[10pt, a4paper]{article}

\usepackage[final]{lrec2026} 
\usepackage{amsmath}
\usepackage{graphicx}

\title{\textbf{\textbf{Building an Ensemble LLM Semantic Tagger for UN Security Council Resolutions}
 } }

\name{Hussein Ghaly} 

\address{Independent Researcher \\
         Rego Park, New York\\
         hmghaly@gmail.com\\}

\abstract{
This paper introduces a new methodology for using LLM-based systems  for accurate and efficient semantic tagging of UN Security Council resolutions. The main goal is to leverage LLM performance variability to build ensemble systems for data cleaning and semantic tagging tasks. We introduce two evaluation metrics: Content Preservation Ratio (CPR) and Tag Well-Formedness (TWF), in order to avoid hallucinations  and unnecessary additions or omissions to the input text beyond the task requirement. These metrics allow the selection of the best output from multiple runs of several GPT models.  GPT-4.1 achieved the highest metrics for both tasks (Cleaning: CPR 84.9\% - Semantic Tagging: CPR  99.99\% and TWF 99.92\%). In terms of cost, smaller models, such as GPT-4.1-mini, achieved comparable performance to the best model in each task at only 20\% of the cost. .  These metrics ultimately allowed the ensemble to select the optimal output (both cleaned and tagged content) for all the LLM models involved, across multiple runs. With this ensemble design and the use of metrics, we create a reliable LLM system for performing semantic tagging on challenging texts.   \newline\newline   }

\begin{document}

\maketitleabstract

\section{Introduction}

Semantic tagging of documents has long been an important goal for enabling the construction of knowledge graphs, and for improving the structure of textual data and making documents more machine-readable. 

Over the past decades, several standards, ontologies, and approaches have been proposed as theoretical frameworks for the logical organization of concepts and entities. 

A lot of work has been performed in the area of semantic tagging and the development of knowledge graphs and other knowledge representations since the early 2000s. Much of this work falls under information extraction, classification, and other natural language processing tasks applied to unstructured text.

For the United Nations, knowledge graphs and machine readability are of paramount importance, given the massive documentation by  the organization in a variety of fields, especially in the areas of maintaining peace and security, human rights, and sustainable development. Many entities and programs have overlapping mandates to serve these goals, and hence there is a critical need to have a meaningful structure of documentation. There have been some initiatives to develop machine readability standards more tailored to UN documentation, mainly using Akoma Ntoso \cite{palmirani_akoma-ntoso_2011}, and UN ontology \footnote{https://unsceb-hlcm.github.io/onto-undo/ }. 

Some limited work has been performed to automatically mark up UN documents with semantic  tags and metadata and for building knowledge graphs \cite{rodven-eide_unsc-graph_2023}. 

With the availability of LLMs since the early 2020s, there is an opportunity for using such systems for the task of semantic tagging, which is the focus of this current paper. 

As we can see from the recent literature, there have been attempts to use LLMs for named entity recognition (NER) \cite{wang_gpt-ner_2025} and for tagging  legal documents \cite{liga_fine-tuning_2023}. There are several such modern approaches, using both Pretrained Language Models and Large Language Models, as shown in a recent survey paper \cite{zhang_survey_2025}.

LLMs have the capability to do the tagging with zero-shot, just by using the right prompt to get the system to tag any input text in a certain way. However, there are many inherent challenges.

The main challenge is variable performance. It is understood that LLMs are stochastic models, where the generation of words and tokens are dependent on different statistical parameters, and hence the output can change every time even with the same input. The main factor controlling this variability is the temperature parameter. In OpenAI models and APIs, it is a setting that varies from 0 to 2. If it is zero, the model consistently gives the same output to the same input, if it is 2, it gives wildly different outputs. Having the same consistent answer can be needed in some applications, but having multiple different possibilities can also be useful, so this variability and temperature parameter are important design considerations in the development of any LLM-based system.

The other main challenge is measurability: How to determine if an output of LLM is good enough or which model provides the better answer for certain tasks or data. 

These challenges affect how to build a system for semantic tagging. For example, we do not want to have a system that over-generates or hallucinates text other than the text that needs to be tagged. Also, we do not want it to under-generate, and skip parts of what needs to be tagged. It needs to be faithful to preserve the existing input content to be tagged, while adding only the tags and necessary requested modifications.

Other factors include ensuring that the tag structure is consistent and well formed. Therefore, additional measures should be introduced to ensure that tags are placed correctly, and each open tag is followed properly by a closing tag.

There are other considerations that relate to the model selection, including model speed and cost, in order to choose a model that performs the semantic tagging task accurately and efficiently, with the most optimum time and cost requirements.

\paragraph{Contributions}

This paper presents a practical and scalable methodology for semantic tagging of UN Security Council resolutions using large language models. The main contributions of this work are:
\begin{itemize}
    \item LLM pipeline for cleaning and semantically tagging historical UN Security Council resolutions.
    \item New evaluation metrics for LLM-based document transformation tasks, including \textit{Content Preservation Ratio (CPR)},  and \textit{Tag Well-Formedness (TWF)}.
    \item Empirical comparison of several LLM models of different sizes and generations, analyzing trade-offs between accuracy and cost.
    \item A semantically annotated sample corpus of UN Security Council resolutions that can support downstream applications including knowledge graph construction.
\end{itemize}

\section{\textbf{Data} }
\label{sec:append-how-prod}

The data used in this project are the text files for security council resolutions, spanning the period of 1946-2025. These documents are part of the project: Corpus of Resolutions: UN Security Council (CR-UNSC)\footnote{https://zenodo.org/records/15154519 } \cite{fobbe_corpus_2024}.

These consist of 2798 files in English. It is important to note that early resolutions, mostly prior to 2000, are produced by typewriters, and they are scanned and their text was produced with OCR systems. Also, it is important to note that until  the 1980s, these documents were produced in two column format, where sometimes the second column is the French translation of the English text in the first column, as in figure X. This two column format is particularly difficult to process with typical NLP tools, since one line of the file can have two chunks of text from different parts of the document (Figure 1). 
\begin{figure}
    \centering
    \includegraphics[width=1\linewidth]{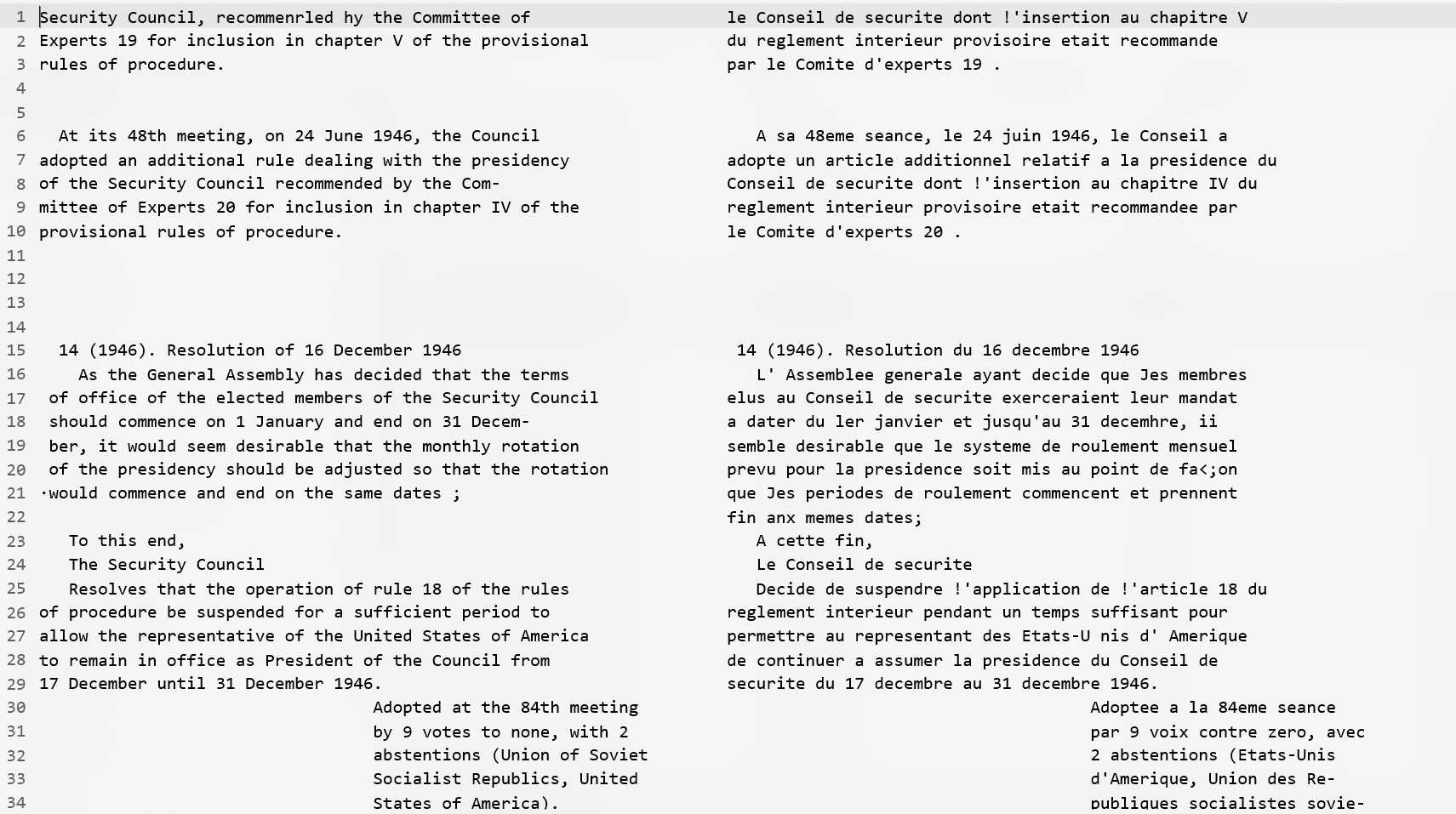}
    \caption{example of noisy 2-column format of early Security Council resolutions }
    \label{fig:placeholder}
\end{figure}

\section{\textbf{Experimental Setup} }

The AI tools for this project are mainly those provided by OpenAI API, which reflect essentially the same capabilities as different versions of ChatGPT, or GPT models in general.

\subsection{\textbf{Prompts}}

For the cleaning part, we use this prompt:

“Include answer only: this text is scanned old UN resolution that can be in two columns, can you convert it into one column and correct any OCR errors, and remove printing hyphens and printing line breaks - if there is English and French texts, separate them, and keep the English first and French second”

For the tagging part, we use this prompt:

“Include answer only: use xml tags to annotate this text, highlighting text within the ONLY these tags, while completely preserving the input text without any addition or omission: 

<location> <entity> <event> <organization> <date>”

No variations of these prompts were used, in order to ensure consistency. However, it is possible to design additional experiments with multiple prompt variations, possibly adding more controls for the desired output.

\subsection{\textbf{Request Parameters}}
For any request to the OpenAI API, we address the following parameters. 
\subsubsection{Temperature}
The variability of model output based on  its input (prompt + input content to be cleaned and tagged) can be a challenge. However, it can also be an opportunity, allowing us to choose the output that is most faithful, consistent, and which best preserves the input content. Therefore, we use a temperature of 1, and have multiple runs, while choosing the best output based on content preservation and other metrics.

\subsubsection{Tokens}

Running the request to OpenAI API requires specifying the maximum number of tokens in the request. We use maximum tokens of 8000, and can vary this parameter depending on the size of the content to be processed.

Token consumption for each request is additionally reported. Each model has a different tokenization scheme for input and output, and this can reflect the cost of running the model\footnote{https://developers.openai.com/api/docs/pricing }.

\subsubsection{Response format}

For many information extraction tasks, the appropriate response would be json format. However, for this particular task, it is most appropriate to use simple text, so no response format needs to be specified.

\subsubsection{Models}

A sample of models have been used, to reflect the diversity among older and newer models (GPT-4 and GPT-5), and also among models of different sizes (full models in addition to mini and nano models). The models are the following:

\begin{itemize}
    \item GPT-4.1
    \item GPT-4.1-mini
    \item GPT-4.1-nano
    \item GPT-4o
    \item GPT-5-mini
    \item GPT-5-nano
    \item GPT-5.1
\end{itemize}

\subsection{\textbf{Metrics}}

To select the best output for each given input, we use certain metrics that reflect the quality of the output for the current tasks. 

\subsubsection{Content Preservation Ratio (CPR)}

For both cleaning and tagging tasks, it is critical to ensure faithfulness, where the output needs to adhere to the input without any deviation. For this purpose, we include a content preservation metric, to compare the input and output. We use Content Preservation Ratio as our main metric. It measures character bigram frequency for any given text. It then compares the bigram frequency in the input text to that in the output text. If the frequency of a certain bigram is the same in both texts, there are no omissions or additions. However, if the bigram has higher frequency in input than output, it reflects omission. If it is higher in output than input, it represents addition. Therefore, we take the combined absolute value of the frequency difference in each bigram as our main measure for content preservation.

 \[
CPR = \frac{\sum_{b} c_{in}(b)}
{\sum_{b} c_{in}(b) - \sum_{b} |c_{in}(b) - c_{out}(b)|}
\]

where \(c_{in}(b)\) is the frequency of a bigram \(b\) in the input text, and \(c_{out}(b)\) is the frequency of the bigram in the output text.

We use bigram frequency, as opposed to minimum edit distance, for example, because of the noisy input due to the two-column format.

\subsubsection{Tag Well-formedness (TWF) }

To perform any tagging task with a makeup language (XML, HTML, or other), it is crucial to ensure that every open tag (<tag>) has a corresponding closing tag (</tag>), and vice versa. For this, we implement a simple algorithm that tokenizes tagged text while preserving the tags as separate tokens. We keep track of all open tags, and if a closing tag is encountered, it needs to match the last open tag, if not (e.g. <tag1><tag2></tag3>), then it is not well-formed. Also, we need to ensure that, by the end of the tagged text, every open tag was closed by the corresponding closing tag and no open tags remain. We calculate all the correctly closed tag pairs, and also those that are not well formed. We use a ratio for well-formedness as the following

\[
TWF = \frac{N_{\text{pairs}}}{N_{\text{pairs}} + N_{\text{malformed}}}
\]

\subsubsection{Number of tags found (nT)}

For the semantic tagging task, it is essential to identify as many tags as possible. At this stage, there is no human evaluation or gold standard dataset against which to measure the accuracy of any of the tags identified. Therefore, the main goal is to optimize for recall: building systems that can identify as many possible tags, even with the risk of having false positives or spurious tags. In fact, further stages of building an ensemble of LLM taggers can allow for more confidence when there is more agreement among the outputs of multiple models or runs regarding certain tags. However, for now we consider the output of the tagger to be better when it can identify more tags.

\begin{figure*}[h!]
    \centering
    \includegraphics[width=1\linewidth]{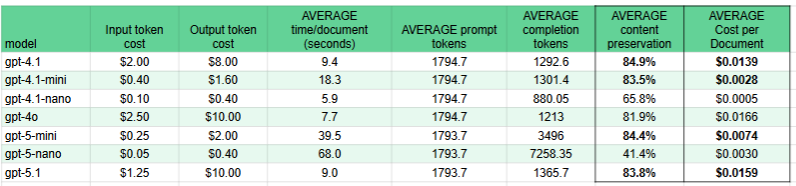}
    \caption{results for the cleaning task }
    \label{fig:placeholder2}
\end{figure*}

\begin{figure*}
     \centering
     \includegraphics[width=1\linewidth]{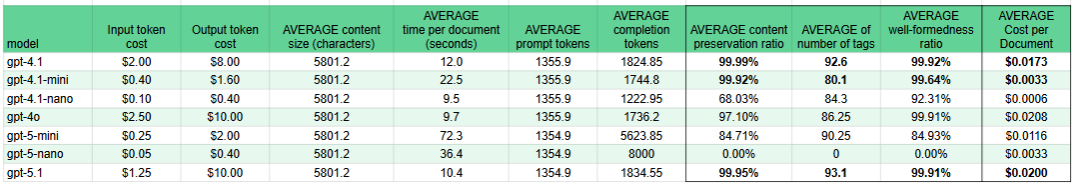}
     \caption{results for the semantic tagging task }
     \label{fig:placeholder3}
 \end{figure*}

\section{\textbf{Methodology} }

Possibly the best utilization of LLMs is when there is a requirement for complex text processing and manipulation that cannot be performed by usual NLP techniques. In our case here, processing old UN Security council resolutions presents an enormous challenge. These documents are written by typewriters, scanned, and then their text was retrieved by Optical Character Recognition (OCR) methods, involving typographical errors. Also the two column format is very challenging, especially in early years, where both the English and French texts were included as opposing columns. Therefore, in order to proceed with any tagging effort, it is essential to have the document in a clean, single column format, with line breaks corresponding to ends of sentences and paragraphs. 

Therefore, we proceed in two tasks:

\begin{itemize}
    \item Cleaning the text of documents
    \item Applying semantic tags into to cleaned text
\end{itemize}

\subsection{Cleaning}

To get the LLM to do the cleaning, we use the prompt indicated in the prompts section above, combined with the raw text content of the resolution to create a full input prompt. We specify the parameters for running the request, including a temperature of 1 to allow variability of output. 

We run the full input prompt for a selection of seven models, listed above.

For each document, we apply the same prompt for each model multiple times (for the current experiment we used only 2 runs per model). 

For each of the 14 outputs (7 models x 2 runs) for the current document, we measure the content preservation ratio (CPR), and identify which output has the highest value of this metric. This output with the highest CPR will be the final output of the LLM ensemble for this document, and the cleaned document will be saved to the data repository.

In addition, we keep track of all the data related to each output, in terms of the model used, number of tokens consumed in the input and output, processing time, and the value of the content preservation ratio for this output. 

\subsection{Semantic Tagging}

Starting from a list of cleaned documents, we proceed to the task of applying the semantic tags. With the prompt indicated above, the tags we focus on are the following, together with their closing counterparts:

<location> <entity> <event> <organization> <date>

Similar to the cleaning part, we also apply each of the 7 GPT models twice on each document, and report the relevant metrics. In this task, the metrics are: content preservation ratio (CPR), tag well-formedness (TWF) and number of tags identified (nT). 

The ensemble sorts each output first according to the content preservation ratio, then well-formedness, then number of tags identified. The best output, according to this sorting, is saved into the repository of tagged documents.

In addition, the performance of each model is measured for each run, to allow for better analysis for all the metrics involved.

\section{\textbf{Results and Discussion} } 

We ran the LLM system, for both cleaning and semantic tagging, on a sample of 10 documents in order to produce clean, tagged outputs of UN Security Council resolutions dated back from 1946. In addition, we analyzed the performance of LLM models on these documents, against our set of metrics, along with practical considerations, such as cost and processing time.

The produced files are stored, along with the code used, into the GitLab repository for illustration purposes only\footnote{https://gitlab.com/acl2575601/kg-llm-2026 }. For actual production, further fine tuning is possible with the inclusion of more models and runs, along with more controlled prompts.

For analyzing the performance of models, we separate the analysis into each of the two tasks (cleaning task: figure 2, semantic tagging task: figure 3).

For the cleaning task, as we can see in figure 2, the best model in terms of content preservation is GPT-4.1, with a ratio of 84.9\% at a cost of \$0.0139 per document. Other models produce comparable ratios, including some smaller models. This can be reflected in the calculation of the cost of processing each document. So, for example, we find that the model GPT-4-mini produces a ratio of 84.4\%, with a cost of \$0.0074 per document, which is almost 53\% of the model with the highest value of this metric. Also, the model GPT-4.1-mini produces a ratio of 83.5\%, with a cost of \$0.0028 per document, which is almost 20\% of the best model for this task. 

We can also see the variations in the processing time for each document. The smaller nano models are usually faster, but they tend to perform much worse in cleaning documents.

For the semantic tagging task (figure 3), we can see similar trends. The best model in terms of both content preservation ratio (CPR), tag well-formedness (TWF), and number of tags identified (nT) is between GPT-4.1 and GPT-5.1. For GPT-4.1, the metrics are the following: CPR 99.99\%, nT 92.6, TWF 99.92\%,  with the cost of processing a document being \$0.017. For GPT-5.1, the metrics are the following: CPR 99.95\%, nT 93.1, TWF 99.91\%, with a cost of \$0.0200 per document.

Looking into other models, we can also find that the model GPT-4.1-mini achieves comparable metrics, but with lower number of tags found: CPR 99.92\%, nT 80.1, TWF 99.64\%, with a cost of \$0.0033 per document (almost 19\% of the cost of the GPT-4.1 model). 

This again suggests the possibility to measure the performance of LLM systems for different types of tasks and data. These measurements allow us to choose the optimal models for the most accurate, reliable, and cost efficient performance for any given task.  

Finally, the output of the ensemble is saved in a structured format, such as XML, where the tags are both human-readable and machine-readable, as in figure 4.
\begin{figure}
    \centering
    \includegraphics[width=1\linewidth]{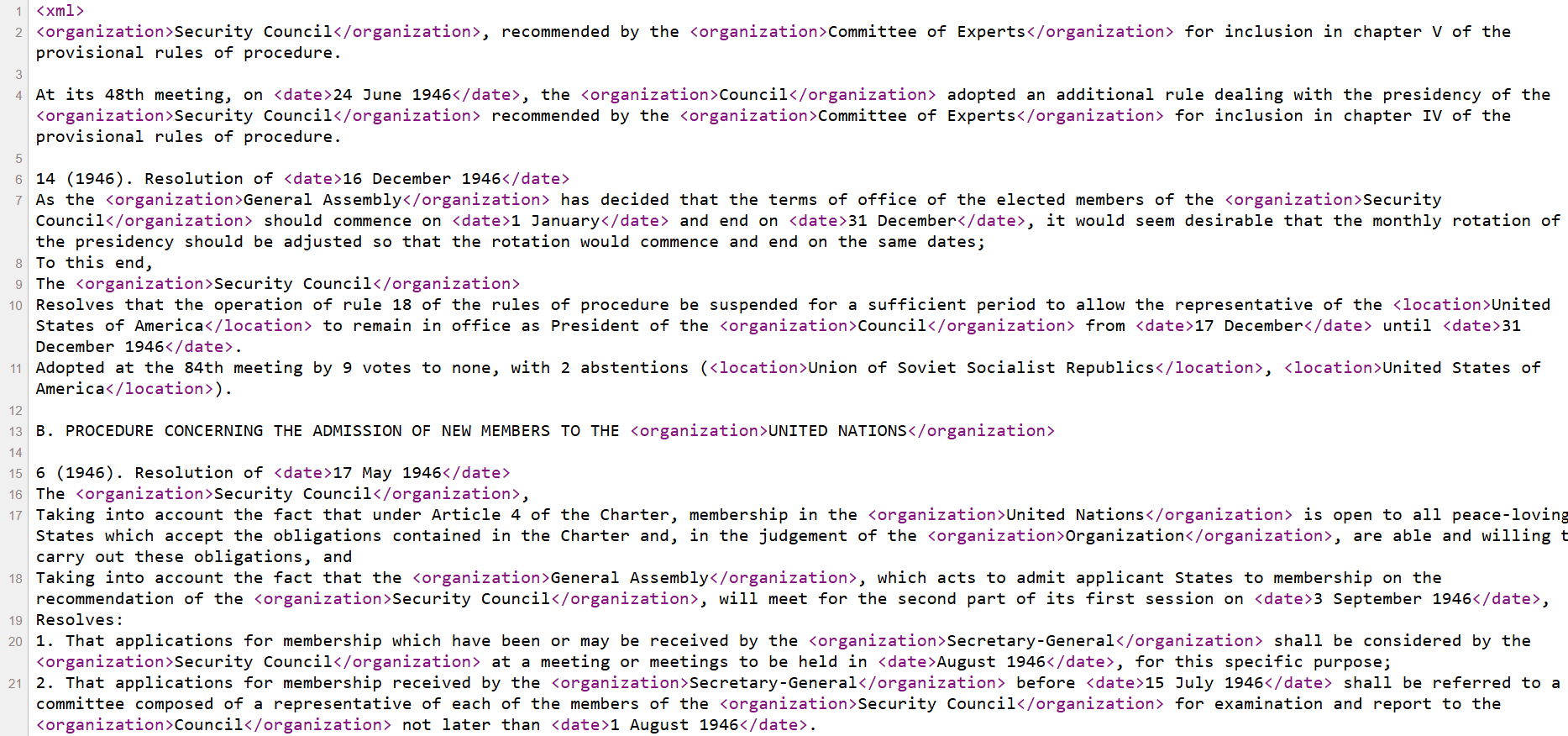}
    \caption{Example of semantic tagging output}
    \label{fig:placeholder4}
\end{figure}
Going forward, the resulting semantically tagged corpus will enable the construction of structured knowledge graphs representing entities, mandates, resolutions,  events, and other substantive aspects, consistent with Akoma Ntoso and other relevant standards, within the UN Security Council archive. 

\section{Conclusion}
Semantic tagging of documents is a first step towards building robust and reliable knowledge graphs. Once it is done properly, it can lead to establishing many connections between entities and concepts from real-world data. 

In this paper, we explored a promising approach for performing semantic tagging on noisy, but rich and relevant content, that of the UN Security Council resolutions, dating back to the 1940s.

Building on the state-of-the-art technologies and capabilities facilitated by LLMs, but with a rigorous approach to ensure proper measurement and control, there is a clear potential for building a full-fledged system for semantic tagging, and hence building knowledge graphs for UN resolutions and possibly other documents.

Also, from a data engineering perspective, it is possible, with this approach, to identify the optimal LLM implementation for a particular task or set of data,. his approach allows for massive commercial savings, especially for data-intensive and big data tasks. In our work, we are able to identify cheaper models that can achieve the desired task with similar performance. Models such as GPT-4.1-mini performed at or below 20\% of the cost of the bigger model, while producing comparable results, for both cleaning and tagging tasks.

Design of an ensemble semantic tagging system requires further analysis. The current ensemble is basically choosing one output, out of several, for any given input, based on the value of certain metrics. In future work we aim to design a  full-fledged ensemble that will  be able to integrate all the outputs together. This is especially relevant for a task such as semantic tagging, where different systems can identify different tags across the input document. Therefore, this multitude of outputs and their tags allows more confidence in identifying the tags agreed upon in most outputs, and eventually combining all of them, in one consolidated and well-formed output. Another important next step is to build the semantic tagger more methodologically according to existing standards, such as Akoma Ntoso XML schemas, allowing the full scope of machine readability of UN documents, and building knowledge graphs accordingly.

This work additionally provides a proof of concept  for harnessing LLMs in a more controlled and measurable way towards annotating unstructured input and building datasets that would be very costly and almost impossible to build manually. These datasets can be used as training data for more Machine Learning and NLP systems, and for establishing more transparent and efficient knowledge systems, facilitated by AI.

\section{Ethics Statement}
This work attempts to have a more transparent and predictable use of AI systems, allowing better controls, constraints and structure. It would allow humans to work better with the outputs of AI systems.

\section{Acknowledgements}

\section{References}\label{sec:reference}

\bibliographystyle{lrec2026-natbib}

\bibliography{LREC2026/references}

\section{Language Resource References}
\label{lr:ref}
\bibliographystylelanguageresource{lrec2026-natbib}
\bibliographylanguageresource{languageresource}

\end{document}